\documentclass[10pt,twocolumn,letterpaper]{article}

\usepackage{iccv}
\usepackage{times}
\usepackage{epsfig}
\usepackage{graphicx}
\usepackage{amsmath}
\usepackage{amssymb}
\usepackage{booktabs}
\usepackage{xcolor}         
\usepackage{colortbl}
\usepackage{multirow}
\usepackage{multicol}
\usepackage{pifont} 
\usepackage{caption}
\usepackage{subcaption}
\usepackage[pagebackref=true,breaklinks=true,letterpaper=true,colorlinks,bookmarks=false]{hyperref}

\usepackage[accsupp]{axessibility}

\newcommand{\cmark}{\ding{51}}%
\newcommand{\xmark}{\ding{55}}%
\newlength\savewidth\newcommand\shline{\noalign{\global\savewidth\arrayrulewidth
  \global\arrayrulewidth 1pt}\hline\noalign{\global\arrayrulewidth\savewidth}}
  
\newcommand{\tablestyle}[2]{\setlength{\tabcolsep}{#1}\renewcommand{\arraystretch}{#2}\centering\footnotesize}

\usepackage[capitalize]{cleveref}
\crefname{section}{Sec.}{Secs.}
\Crefname{section}{Section}{Sections}
\Crefname{table}{Table}{Tables}
\crefname{table}{Tab.}{Tabs.}

\iccvfinalcopy 

\def\iccvPaperID{5484} 

\ificcvfinal\fi

\begin{document}

\title{Dynamic Perceiver for Efficient Visual Recognition}

\author{Yizeng Han$^1$\thanks{Equal contribution.}~~Dongchen Han$^{1*}$~Zeyu Liu$^{2}$~Yulin Wang$^{1}$~Xuran Pan$^{1}$~Yifan Pu$^{1}$\\
Chao Deng$^{3}$~Junlan Feng$^{3}$~Shiji Song$^{1}$~Gao Huang$^{1,4}\thanks{Corresponding Author.}$ \\ 
{\small $^1$ Department of Automation, BNRist, Tsinghua University}\\
{\small $^2$ Department of Computer Science and Technology, BNRist, Tsinghua University}\\
{\small $^3$ China Mobile Research Institute~~~~~$^4$ Beijing Academy of Artificial Intelligence}\\
}

\maketitle
\ificcvfinal\thispagestyle{empty}\fi

\begin{abstract}
   Early exiting has become a promising approach to improving the inference efficiency of deep networks. By structuring models with multiple classifiers (exits), predictions for ``easy'' samples can be generated at earlier exits, negating the need for executing deeper layers. Current multi-exit networks typically implement linear classifiers at intermediate layers, compelling low-level features to encapsulate high-level semantics. This sub-optimal design invariably undermines the performance of later exits. In this paper, we propose \textbf{Dynamic Perceiver (Dyn-Perceiver)} to decouple the feature extraction procedure and the early classification task with a novel dual-branch architecture. A feature branch serves to extract image features, while a classification branch processes a latent code assigned for classification tasks. Bi-directional cross-attention layers are established to progressively fuse the information of both branches. Early exits are placed exclusively within the classification branch, thus eliminating the need for linear separability in low-level features. Dyn-Perceiver constitutes a versatile and adaptable framework that can be built upon various architectures. Experiments on image classification, action recognition, and object detection demonstrate that our method significantly improves the inference efficiency of different backbones, outperforming numerous competitive approaches across a broad range of computational budgets.  Evaluation on both CPU and GPU platforms substantiate the superior practical efficiency of Dyn-Perceiver. Code is available at \url{https://www.github.com/LeapLabTHU/Dynamic_Perceiver}.
\end{abstract}
\section{Introduction} \label{sec:intro}
\begin{figure}
    \begin{center}
      \includegraphics[width=\linewidth]{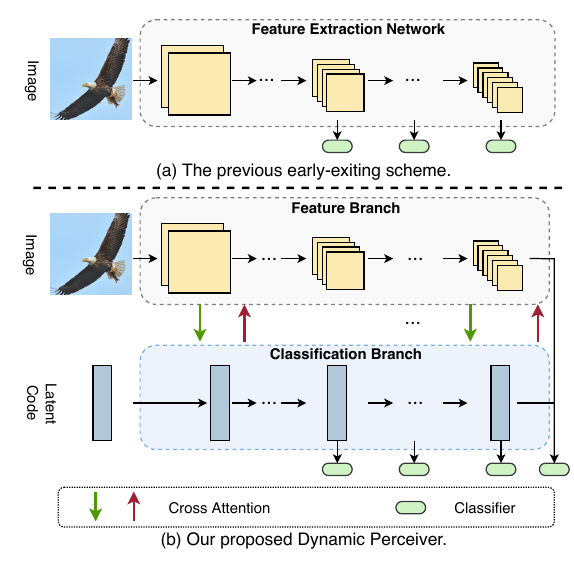}
    \end{center}
    \vskip -0.25in
    \caption{Comparison of Dyn-Perceiver with the previous early-exiting scheme. (a) Conventional methods build classifiers on \emph{intermediate features}, degrading the performance of the last exit; (b) Dyn-Perceiver \emph{decouples} feature extraction and early classification with a \emph{two-branch} structure.}
    \label{fig1_idea}
    \vskip -0.25in
\end{figure}
Convolutional neural networks (CNNs) \cite{he2016resnet,huang2017densely,xie2017aggregated, radosavovic2020designing,liu2022convnet,lin2019tsm} and vision Transformers  \cite{dosovitskiy2020image,touvron2021training,yuan2021tokens,liu2021swin,wang2021pyramid,liu2022video} have precipitated substantial advancements in visual recognition. Despite concerted efforts towards scaling up vision models for superior accuracy  \cite{zhai2022scaling,liu2022swin,riquelme2021scaling}, the high computational demands have acted as a deterrent to their deployment in resource-constrained scenarios.  
Research endeavours towards improving the inference efficiency of deep networks span a multitude of directions, including
lightweight architecture design \cite{howard2017mobilenets,zhang2018shufflenet,howard2019searching}, pruning \cite{han2015deepcompression,he2018pruning,yang2021condensenet}, quantization \cite{jung2019learning,yuan2022ptq4vit}, {etc.} 
In contrast to traditional models, which adhere to a \emph{static} computational graph during testing, \emph{dynamic networks} \cite{han2021dynamic,bolukbasi2017adaptive,lin2017runtime,wang2021adaptive,han2022latency,wang2022adafocus,wang2022adafocusv3,yl_spatial_video,zhao2023ada3d} can adapt their computation with varying input complexities, leading to promising results in efficient visual recognition. 

In the field of dynamic networks, dynamic \emph{early-exiting} networks \cite{bolukbasi2017adaptive,graves2016adaptive,figurnov2017spatially,huang2017multi,yang2020resolution,adadet}  construct multiple classifiers along the depth dimension, allowing samples that yield high classification confidence at early classifiers, referred to as ``easy'' samples, to be rapidly predicted without activating deeper layers.
Existing implementations mostly build early classifiers on intermediate features \cite{bolukbasi2017adaptive,huang2017multi,yang2020resolution} (\cref{fig1_idea} (a)). However, it has been observed \cite{huang2017multi} that classifiers will interfere with each other and significantly degrade the performance of the final exit. A widely held belief is that deep models generally extract features from a low level to a high level, and it is more appropriate to feed the high-level features at the \emph{end} of a network to a linear classifier. Early classifiers in previous literature force \emph{intermediate} low-level features to encapsulate high-level semantics and be linearly separable. This essentially means that \emph{feature extraction} and \emph{early classification} are \emph{intricately intertwined}. This sub-optimal design invariably undermines the performance of dynamic early-exiting networks.

Ideally, it is expected that 1) there is a \emph{latent code} which consistently embeds semantic information for direct use in classification tasks;
2) early classification and feature extraction should be \emph{decoupled}, i.e., the acquisition of semantic information should be managed by a separate branch, thereby avoiding the necessity of sharing shallow layers in a feature extractor.
Under these circumstances, the latent code needs to achieve linear separability, not the low-level image features, preserving the performance of late exits.
The concept of incorporating a latent code is inspired by the general-purpose architecture, Perceiver \cite{pmlr-v139-jaegle21a}. This model leverages  \emph{asymmetric} attention to iteratively \emph{distill inputs into a latent code}, which is then employed for specific tasks. Despite its impressive ability to process various modalities, Perceiver's application in visual recognition encounters a significant challenge in terms of computational cost, particularly when the pixel count in images is substantial.

In this paper, we propose a novel two-branch structure (\cref{fig1_idea} (b)), named \textbf{Dynamic Perceiver (Dyn-Perceiver)}, for efficient visual recognition. Specifically, a \emph{feature branch} extracts image features from a low level to a high level. Concurrently, 
a \emph{trainable latent code}, engineered to encapsulate the semantics pertinent to classification, is processed by a \emph{classification branch}.
These two branches progressively exchange information via \emph{symmetric} cross-attention layers,
and the token number of image features is significantly reduced compared to the original Perceiver. Critically, 
multiple classifiers are situated \emph{solely in the classification branch}, enabling early predictions without hindering feature extraction. The outputs from both branches are ultimately fused before being supplied to the final classifier.

Our design boasts three key advantages:
1) feature extraction and early classification are \emph{explicitly decoupled}, and the experiment results in \cref{sec:ablation} demonstrate that the early classifier in our method even \emph{improves} the performance of the last exit; 2) the Dyn-Perceiver framework is simple and versatile.
It does away with the need for meticulously handcrafted structures as seen in previous approaches 
\cite{huang2017multi,yang2020resolution,wang2021images}. In essence, we can construct the classification branch on any advanced vision backbones to attain top-tier performance. Such universality also allows Dyn-Perceiver to seamlessly serve as a backbone for downstream tasks such as object detection; 3) the theoretical efficiency of early exiting in Dyn-Perceiver can effectively translate into practical speedup on different hardware devices.

We evaluate the performance of Dyn-Perceiver with multiple visual backbones including ResNet \cite{he2016resnet}, RegNet-Y \cite{radosavovic2020designing}, and MobileNet-v3 \cite{howard2019searching}. Experiments show that Dyn-Perceiver significantly outperforms various competing models in terms of the accuracy-efficiency trade-off in ImageNet \cite{deng2009imagenet} classification. Notably, 
the inference efficiency of RegNet-Y experiences a remarkable increase of 1.9-4.8$\times$ without any compromise in accuracy. 
The practical latency of Dyn-Perceiver is also validated on CPU and GPU platforms.  Additionally, our method effectively enhances the performance-efficiency trade-off in action recognition (Something-Something V1 \cite{goyal2017something}) and object detection tasks. For instance, Dyn-Perceiver boosts the mean average precision (mAP) of RegNet-Y by 0.9\% while diminishing its computation by 43\% on the COCO \cite{lin2014microsoft} dataset.

\section{Related Works} \label{sec:related_works}

\begin{figure*}
    \begin{center}
      \includegraphics[width=0.95\linewidth]{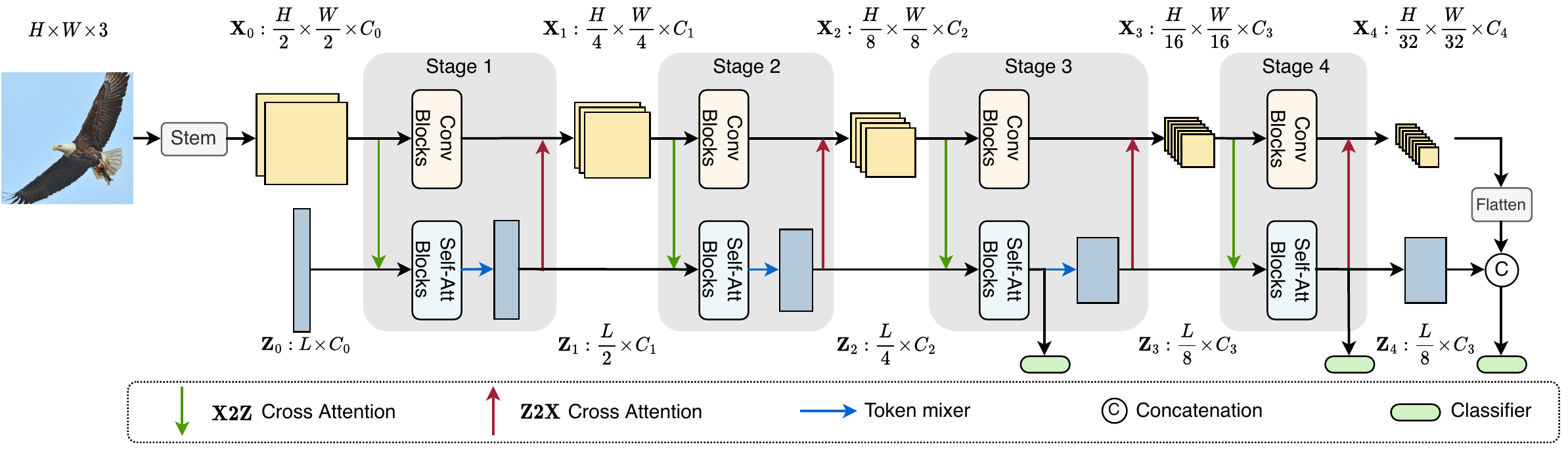}
    \end{center}
    \vskip -0.25in
    \caption{An overview of Dyn-Perceiver. The \emph{feature branch} (top) and the \emph{classification branch} (bottom) process image features $\mathbf{X}_0,\!\cdots,\!\mathbf{X}_4$ and the latent code $\mathbf{Z}_0,\!\cdots,\!\mathbf{Z}_4$ respectively. Cross-attention layers are built symmetrically to fuse information from the two branches. Early classifiers are appended only in the classification branch. Best viewed in color.}
    \vskip -0.15in
    \label{fig2_arch}
\end{figure*}

\textbf{Efficient visual recognition.} Extensive efforts have been dedicated to improving the inference efficiency of deep networks. Popular approaches include network pruning \cite{han2015deepcompression,he2018pruning,yang2021condensenet,tang2022patch}, weight quantization \cite{jung2019learning,hubara2016binarized,yuan2022ptq4vit,liu2021post}, and lightweight architecture design \cite{howard2017mobilenets,zhang2018shufflenet}. However, an inherent limitation of these \emph{static} models is that they all treat different samples with equal computation, leading to inevitable inefficiency. In contrast, our Dyn-Perceiver can \emph{adapt} its architecture (depth) to different inputs, effectively mitigating superfluous computation for ``easy'' samples.

\textbf{Perceiver-style architectures.} Our work draws inspiration from the general-purpose model Perceiver \cite{pmlr-v139-jaegle21a,jaegle2021perceiver}. Perceiver's latent code directly queries information from raw inputs. Such generality comes at the cost of expensive computation. To this end, we adopt a visual backbone as feature extractor, thereby allowing the latent code to efficiently collect information from \emph{features}, which contain considerably fewer tokens. Moreover, the performance of Dyn-Perceiver profits from our \emph{symmetric} attention mechanism. Finally, Perceiver is a \emph{static} model, which recursively executes attention layers for a \emph{fixed} number of times. Dyn-Perceiver \emph{dynamically} skips the computation of deep layers.

Mobile-Former \cite{chen2022mobile} also explores convolution-attention interactions. Dyn-Perceiver differs from Mobile-Former in two key aspects: 1) while Mobile-Former strives to construct an efficient \emph{static} network, our model is a universal \emph{framework} especially designed for \emph{dynamic early exiting}; 2) the convolution's input in a Mobile-Former block is the output from attention, rendering the inference pipeline a \emph{sequential} process. Nevertheless, our computation in two branches is \emph{independent} and hence more parallel-friendly.

\textbf{Dynamic early exiting} \cite{bolukbasi2017adaptive,huang2017multi,yang2020resolution} facilitates swift output predictions at shallower layers, reducing redundant computation in deep layers. Past observations \cite{huang2017multi} have noted that the direct insertion of early classifiers degrades the performance of the final exit. Multi-scale dense network (MSDNet) \cite{huang2017multi} and resolution adaptive network (RANet) \cite{yang2020resolution} \emph{partially} address this via multi-scale structures and dense connections. However, their early classifiers are still appended on intermediate features. As a countermeasure, our model \emph{explicitly decouples} feature extraction and early classification via a dual-branch architecture, which effectively \emph{improves} the performance of the final exit. Furthermore, Dyn-Perceiver is a \emph{general} and \emph{simple} framework. It can be effortlessly constructed atop various backbones without requiring the intricately-designed architectures such as MSDNet. This adaptability allows Dyn-Perceiver to seamlessly function as a backbone for downstream tasks.

\section{Method} \label{sec:method}
In this section, we first provide an overview of the proposed Dyn-Perceiver (\cref{sec:overview}). Then the main components are explained (\cref{sec_main_components}). We finally introduce the adaptive inference paradigm and the training strategy (\cref{sec_training_inference}).

\subsection{Overview} \label{sec:overview}
\noindent\textbf{Overall architecture.} To explicitly decouple the feature extraction process and the early classification task, we propose a novel two-branch architecture consisting of 4 stages (\cref{fig2_arch}). The first branch, refered to as the \emph{feature branch}, can be designed as any visual backbone. In this paper, we implement it as a CNN for efficiency. The feature branch takes an image as input and generates feature maps ($\mathbf{X}_{0}$ to $\mathbf{X}_{4}$) from a low level to a high level. The second branch, denoted as the \emph{classification branch}, receives a trainable latent code $\mathbf{Z}_{0}$ as input. This latent code is randomly initialized and then processed by a series of self-attention operations. Following the common practice in popular vision models \cite{he2016resnet,huang2017densely,liu2021swin}, we construct a \emph{token mixer} (blue arrows in \cref{fig2_arch}) between every two stages to reduce the token length and expand the hidden dimension of the latent code.

\noindent\textbf{Symmetric cross attention.} To incorporate the semantic information into the latent code, we adopt feature-to-latent ($\mathbf{X2Z}$) cross attention (green arrows in \cref{fig2_arch}) at the start of each stage. Subsequently, the two branches conduct convolution and self-attention operations \emph{independently}. The semantic information in the latent code is then integrated into the feature branch via latent-to-feature ($\mathbf{Z2X}$) cross attention (red arrows in \cref{fig2_arch}) at the end of each stage. 

\noindent\textbf{Dynamic early exiting.} The output from two branches are ultimately merged before being input to a linear classifier. Importantly, intermediate classifiers are added at the end of the last two stages of the \emph{classification branch} to facilitate dynamic early exiting without disrupting feature extraction.

\subsection{Main components}\label{sec_main_components}
\begin{figure}
    \begin{center}
      \includegraphics[width=0.95\linewidth]{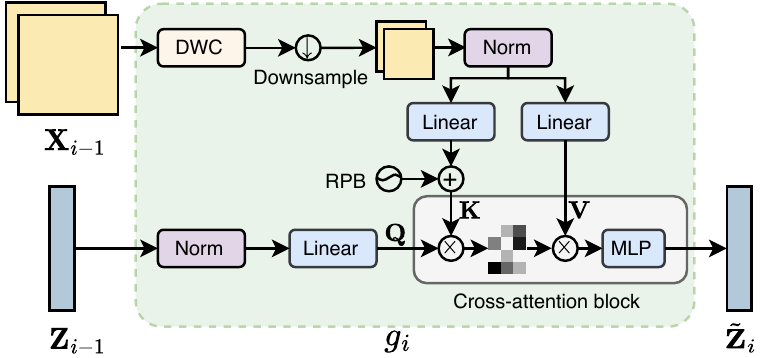}
    \end{center}
    \vskip -0.25in
    \caption{$\mathbf{X2Z}$ \textbf{cross attention}. The operations ($g_i$) in the light green block correspond to a green arrow in \cref{fig2_arch}.}
    \label{fig3_x2z_ca}
    \vskip -0.15in
\end{figure}

In this subsection, we present the main components in Dyn-Perceiver, generally in their order of execution.

\noindent\textbf{$\mathbf{X2Z}$ cross attention.} At the start of each stage, the classification branch interacts with the input by querying information from image features. Specifically, we denote the input of the $i$-th ($i\!=\!1,2,3,4$) stage in the feature branch and the classification branch as $\mathbf{X}_{i-1}$ and $\mathbf{Z}_{i-1}$, respectively. Before feeding $\mathbf{Z}_{i-1}$ to the self-attention blocks, we use a cross-attention module $g_i$ to integrate the image feature $\mathbf{X}_{i-1}$ into the latent code (\cref{fig3_x2z_ca}): $\tilde{\mathbf{Z}}_{i-1}\!=\! g_i(\mathbf{Z}_{i-1}, \mathbf{X}_{i-1}),$
where $\mathbf{Z}_{i-1}$ is the query, and $\mathbf{X}_{i-1}$ is the key and the value. Note that the token numbers in early features can be large, which is inefficient if we directly conduct cross attention. Therefore, we apply depth-wise convolution (DWC) to enhance local feature extraction, and then pool the feature $\mathbf{X}_{i-1}$ to the size of 7$\times$7 before feeding it to the cross-attention block. Moreover, we use relative position bias (RPB) \cite{ramachandran2019stand} to encode the position information.

\noindent\textbf{Self-attention blocks in the classification branch.} The output of $\mathbf{X2Z}$ cross attention $\tilde{\mathbf{Z}}_{i-1}$ is then processed by cascaded self-attention blocks: $\mathbf{Z}_{i}'\!=\!f_i^{\mathrm{att}}(\tilde{\mathbf{Z}}_{i-1}).$ In this paper, we simply implement $f_i^{\mathrm{att}}$ with the standard Transformer \cite{vaswani2017attention} blocks. Specifically, a Transformer block is composed of a multi-head self-attention (MHSA) block followed by a multi-layer perceptron (MLP).

\begin{figure}
    \begin{center}
      \includegraphics[width=\linewidth]{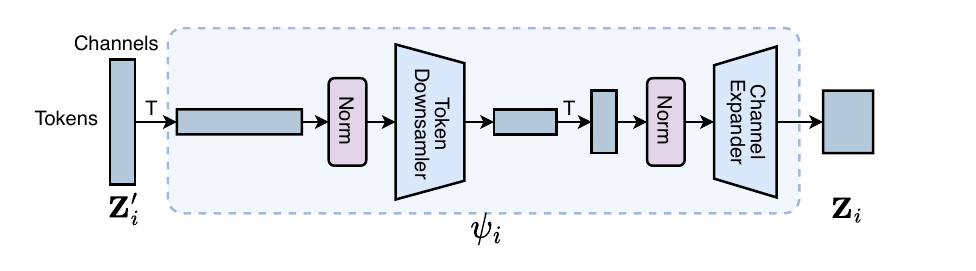}
    \end{center}
    \vskip -0.25in
    \caption{\textbf{Token mixer}. The token downsampler performs per-channel interaction along the token dimension, and the channel expander performs per-token interaction along the channel dimension. The operations in the light blue block ($\psi_i$) correspond to a blue arrow in \cref{fig2_arch}.}
    \label{fig4_token_mixer}
    \vskip -0.2in
\end{figure}

\noindent\textbf{Token mixers.} Inspired by the common practice in popular vision backbones \cite{he2016resnet,huang2017densely,liu2021swin}, we propose to ``downsample'' the latent code and align its channels with the image feature between every two stages. Specifically, we use two linear layers to reduce the token number of the latent code and expand the channel number\footnote{We find in our experiments that the order of these two operations has a neglectable influence on the model performance.} (see \cref{fig4_token_mixer}). We denote the token mixer as $\psi_i$. It takes $\mathbf{Z}_i'$ calculated by $f_i^{\mathrm{att}}$ as input, and generates the output of the $i$-th stage in the classification branch: $\mathbf{Z}_i\! =\! \psi_i(\mathbf{Z}_i')$. To summarize, the operations in stage $i$ of the classification branch can be written as
\begin{equation}\label{eq:z_i}
    \mathbf{Z}_i = \psi_i(f_i^{\mathrm{att}}(g_i(\mathbf{Z}_{i-1}, \mathbf{X}_{i-1}))).
\end{equation}

\noindent\textbf{The feature branch} can be established with an arbitrary visual backbone. In this paper, we primarily experiment with three commonly used CNNs: ResNet \cite{he2016resnet}, RegNet-Y \cite{radosavovic2020designing} and MobileNet-v3 \cite{howard2019searching}. We denote the output of the $i$-th CNN stage as 
$\tilde{\mathbf{X}}_i\!=\!f_i^{\mathrm{conv}}(\mathbf{X}_{i-1})$,
where $f_i^{\mathrm{conv}}$ is the sequential convolutional blocks in the $i$-th CNN stage. Note that $f_i^{\mathrm{conv}}$ can be executed \emph{independently} with the self-attention blocks $f_i^{\mathrm{att}}$ in the first two stages to facilitate parallel computation. We execute the two branches \emph{sequentially} in the last two stages to minimize the computation for obtaining early predictions (see \cref{sec_training_inference} for details).


\begin{figure}
    \begin{center}
      \includegraphics[width=0.95\linewidth]{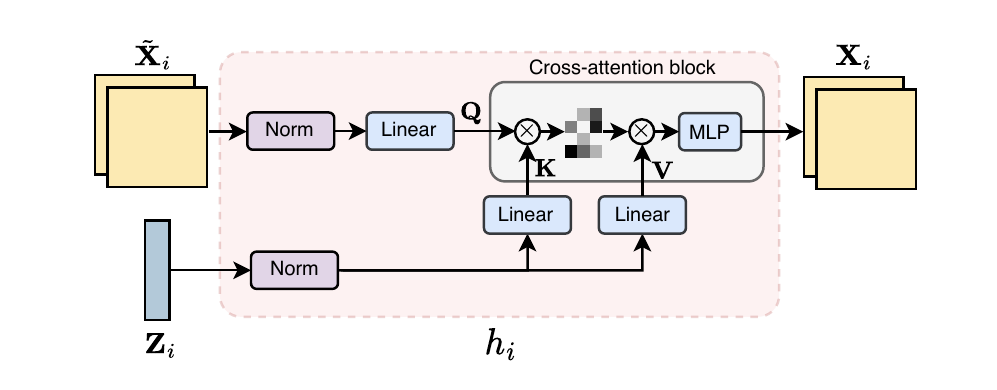}
    \end{center}
    \vskip -0.25in
    \caption{$\mathbf{Z2X}$ \textbf{cross attention}. 
    The operations in the light red block ($h_i$) correspond to a red arrow in \cref{fig2_arch}.}
    \vskip -0.1in
    \label{fig5_z2x_ca}
\end{figure}

\begin{figure}
        \centering
        \includegraphics[width=0.9\linewidth]{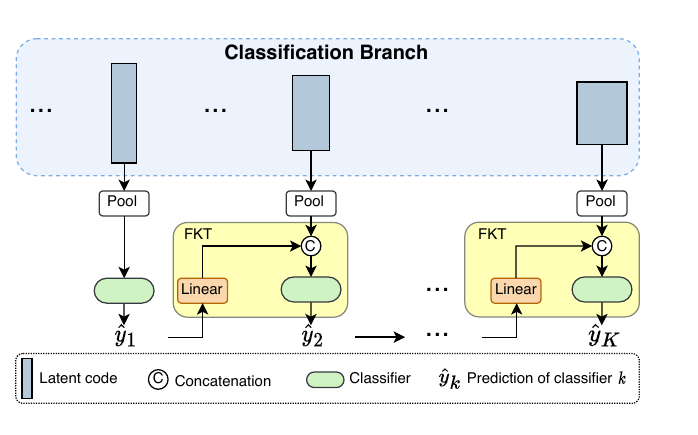}
        \vskip -0.1in
        \caption{FKT modules.}
        \label{fig_fkt}
    \vskip -0.2in
\end{figure}

\noindent$\mathbf{Z2X}$ \textbf{cross attention.} Different from the original Perceiver which only distills the input into the latent code, We further use a cross-attention layer (\cref{fig5_z2x_ca}) at the end of each stage to aggregate the semantic information in the classification branch back into the feature branch. We denote the cross-attention layer after stage $i$ as $h_i$. The output of the convolutional blocks  $\tilde{\mathbf{X}}_i$ is used as the query, and the output of the token mixer $\mathbf{Z}_i$ is the key and the value. The $\mathbf{Z2X}$ cross attention $h_i$ produces the input for the ($i\!+\!1$)-th CNN stage: $\mathbf{X}_{i}\!=\!h_i(\tilde{\mathbf{X}}_i,\mathbf{Z}_i)$. 
In a nutshell, the operations on image features in stage $i$ can be represented by
\begin{equation}
    \mathbf{X}_{i}=h_i(f_i^{\mathrm{conv}}(\mathbf{X}_{i-1}), \mathbf{Z}_i).
\end{equation}

\noindent\textbf{Classifiers.} To conduct dynamic early exiting \cite{han2021dynamic} without disrupting feature extraction, we build classifier heads after the last two stages only in the \emph{classification branch}. Concretely, we pool the latent code along the token dimension and feed the result to a classification head. 
The classifier at the end of the feature branch is kept, as we find it slightly improves the dynamic inference performance. We also find that early classifiers at the first two stages bring limited improvements in dynamic early exiting. Finally, we concatenate the outputs from two branches and establish the last classifier based on the merged features.

\noindent\textbf{Forward Knowledge Transfer (FKT)}. Inspired by the previous work on training multi-exit models \cite{li2019improved}, we propose to transfer the knowledge of early classifiers to deep ones. Specifically, a linear layer is attached to the output of an early classifier. The pooled latent code in the next stage is concatenated with the output of this linear layer before being fed to the classifier (\cref{fig_fkt}). It is worth noting that instead of using a \emph{pretrain-finetune} strategy as in \cite{li2019improved}, our FKT modules can directly improve the performance of both early and deep classifiers in \emph{end-to-end} training (see the empirical analysis in \cref{sec:ablation}). We believe that FKT could be viewed as a shortcut between classifiers, which also facilitates the optimization of early classifiers.

\begin{figure}
    \begin{center}
      \includegraphics[width=0.825\linewidth]{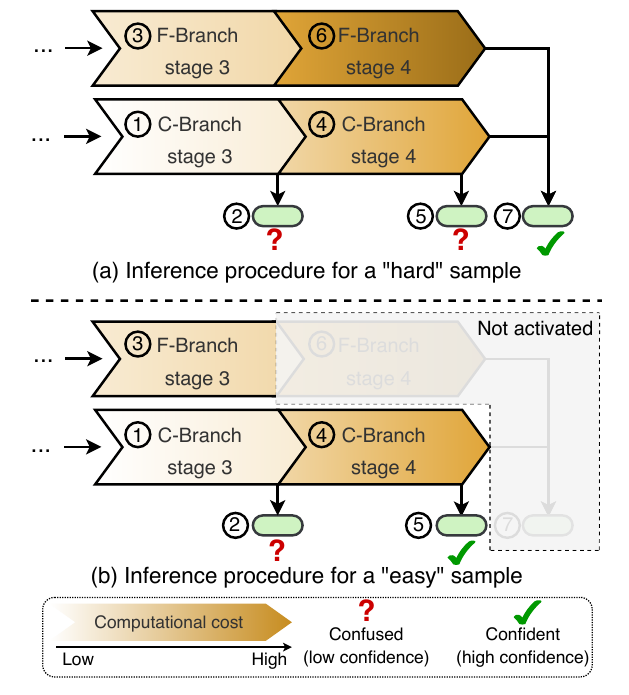}
    \end{center}
    \vskip -0.25in
    \caption{\textbf{The inference procedure} of the last two stages for ``hard'' (a) and ``easy'' (b) samples. F-Branch and C-Branch are the feature branch and the classification branch, respectively. The circled numbers denote the execution order. Deeper layers (shaded areas) will not be activated for the samples that are predicted with high confidence by an early classifier. Best viewed in color.}
    \vskip -0.2in
    \label{fig6_ee}
\end{figure}

\subsection{Inference and training}\label{sec_training_inference}

\noindent\textbf{Dynamic early exiting.} To reduce the redundant computation on ``easy'' samples, we conduct dynamic early exiting based on the classification confidence of early classifiers. The inference procedure for processing ``hard'' and ``easy'' samples is illustrated in \cref{fig6_ee}. We can observe from \cref{eq:z_i} that the output of stage $i$ in the classification branch $\mathbf{Z}_i$ does not rely on the output of the same stage in the feature branch  $\mathbf{X}_i$. Therefore, the early prediction can be obtained by first activating a stage in the  classification branch and its followed classifier. If the confidence (the max value of the \emph{Softmax} probability) exceeds a threshold, the forward propagation terminates without activating deeper layers.

\noindent\textbf{Training with self-distillation.} We propose to use our last classifier to guide the training of early exits. Specifically, the loss function for the $k$-th classifier can be written as 
\begin{align}\label{eq:training}
    \mathcal{L}_k &= \alpha \mathcal{L}_k^{\mathrm{CE}} + (1-\alpha) \mathcal{L}_k^{\mathrm{KD}}, k=1,2,\cdots, K-1, \nonumber \\ 
    \mathcal{L}_K &= \mathcal{L}_K^{\mathrm{CE}},
\end{align}
where $K$ is the number of exits, and  $\mathcal{L}_k^{\mathrm{CE}}$ is the cross-entropy (CE) loss of the $k$-th classifier. The item $\mathcal{L}_k^{\mathrm{KD}}$ is the Kullback Leibler (KL) divergence of the soft class probabilities between the $k$-th and the last classifier. Such self-distillation training is a complementary strategy to the aforementioned forward knowledge transfer (FKT) module and is also used without the pretrain-finetune paradigm in \cite{li2019improved}. The ablation study in \cref{sec:ablation} demonstrates that it can successfully boost the performance of \emph{early} classifiers. 

The overall training loss can be constructed by accumulating the loss from all exits: $\mathcal{L}\!=\!\sum_{k=1}^K \mathcal{L}_k$, and $\alpha$ in \cref{eq:training} is simply set as 0.5 in all our experiments.

\section{Experiments} \label{sec:exp}
In this section, we first evaluate Dyn-Perceiver with different visual backbones on ImageNet \cite{deng2009imagenet}, and then validate the proposed method in action recognition on Something-Something V1 \cite{goyal2017something} (\cref{sec:main_results}). Ablation studies (\cref{sec:ablation}) and visualization (\cref{sec:visualization}) are further presented to give a deeper understanding of our approach. 
Finally, we demonstrate the versatility of Dyn-Perceiver by using it as a backbone for COCO object detection \cite{lin2014microsoft} (\cref{sec:coco_det}).

\begin{figure*}
     \includegraphics[width=\linewidth]{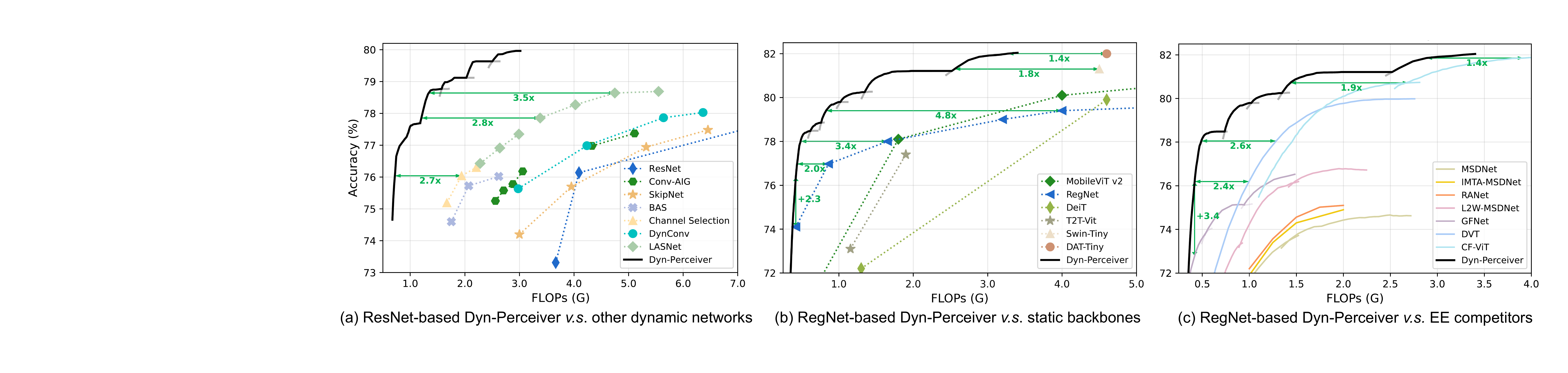}
         \vskip -0.1in
        \caption{Accuracy v.s. FLOPs curves on ImageNet of Dyn-Perceiver implemented on ResNets (a) and RegNets-Y (b, c). Competing dynamic networks, static backbones, and early-exiting methods are compared in (a), (b), and (c) respectively.}
        \label{fig:main_results}
        \vskip -0.1in
\end{figure*}

\begin{figure*}
    \begin{minipage}[t]{0.66\linewidth}
        \centering
        \includegraphics[width=\linewidth]{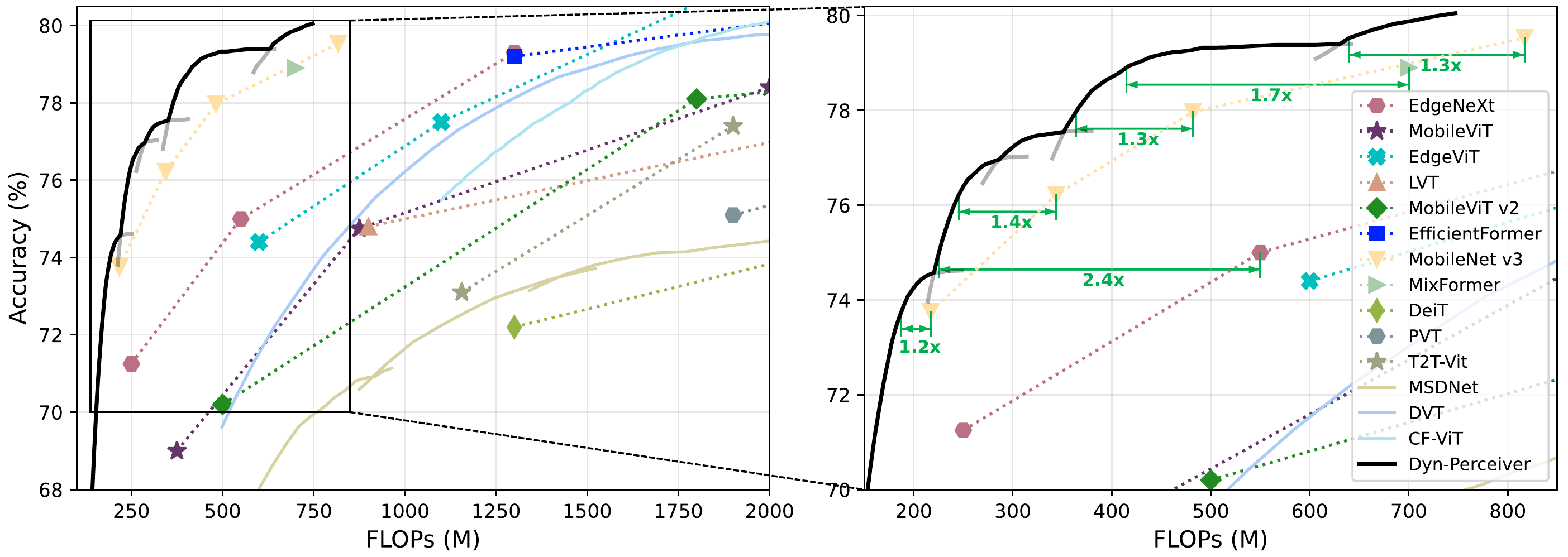}
        \vskip -0.1in
        \caption{ImageNet results of Dyn-Perceiver built on top of MobileNet-v3.}
        \label{fig8_mob_results}
    \end{minipage}\hfill
    \begin{minipage}[t]{0.33\linewidth}
        \centering
        \includegraphics[width=\textwidth]{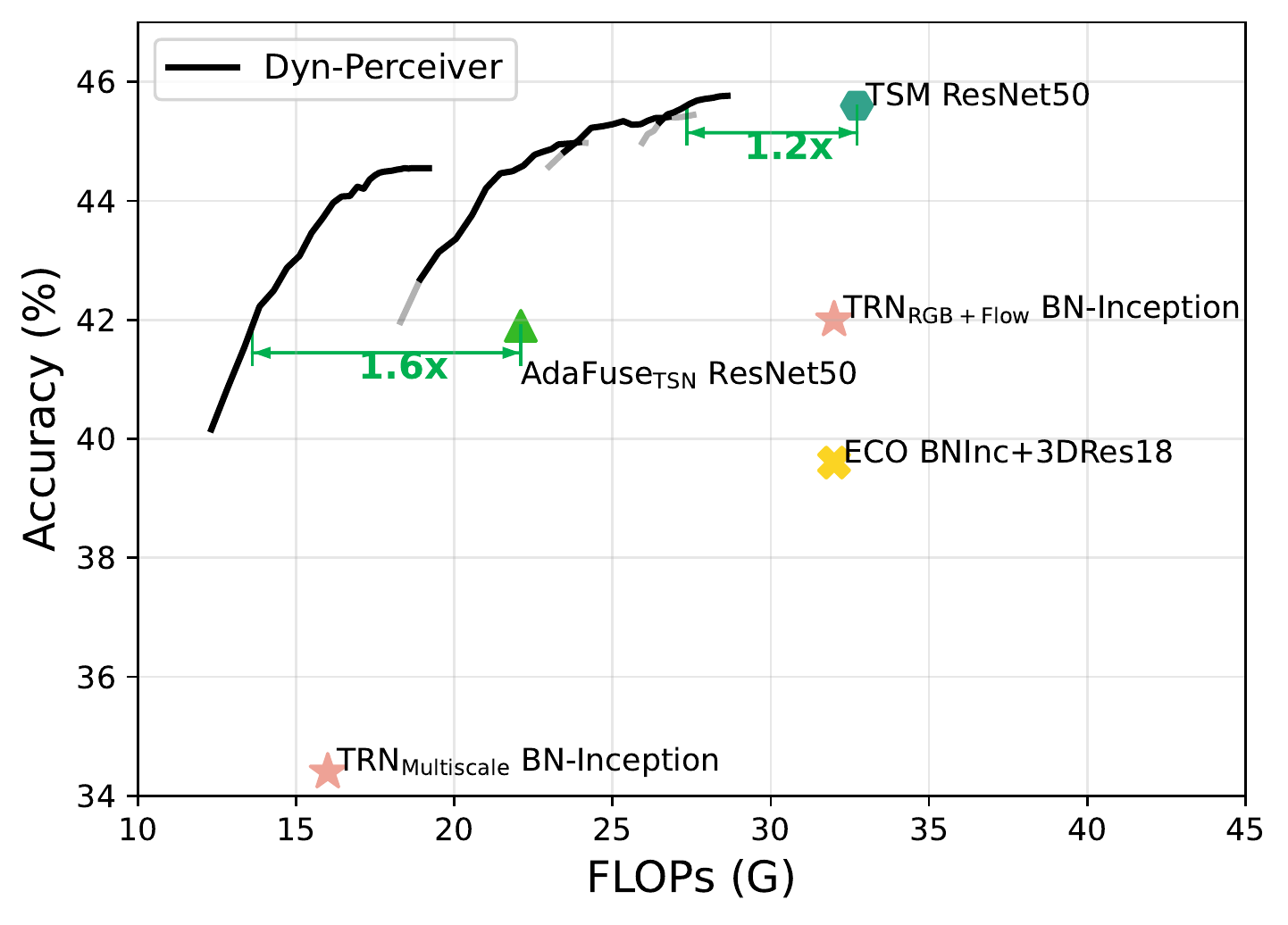}
        \vskip -0.1in
        \caption{Action recognition results.}
        \label{fig_video_results}
    \end{minipage}
    \vskip -0.2in
\end{figure*}

\noindent\textbf{Datasets.} ImageNet \cite{deng2009imagenet} comprises 1000 classes, with 1.2 million and 50,000 images for training and validation. The images in ImageNet are of size 224$\times$224. Something-Something V1 \cite{goyal2017something} is a large-scale human action dataset that includes 98k videos, and we use the official training-validation split. The COCO dataset \cite{lin2014microsoft} contains 80 categories with 118k training images and 5k validation images. We use the average FLOPs (floating-point operations) on the validation set of each dataset to measure the computational cost. The FLOPs are calculated with 8 224$\times$224 frames per video on Something-Something V1 and are computed based on an input size of 1280$\times$800 on COCO.

\noindent\textbf{Models.} We implement the feature branch with ResNet-50 \cite{he2016resnet}, RegNet-Y \cite{radosavovic2020designing} and MobileNet-v3 \cite{howard2019searching}. For the classification branch, we choose the initial token number $L$ of the latent code from \{128,192,256\} to construct different-sized models. The head number of self attention in stage $i$ is fixed as $2^{i-1}, i\!=\!1,\!2,\!3,\!4$. The cross-attention layers all have 1 head. 
Other details are listed in \cref{sec:model_configurations}. 

\noindent\textbf{Inference and training.} 
To perform dynamic early exiting on ImageNet, we randomly split 50,000 images from the training set. Then we vary the computation budget, solve the confidence thresholds on the split data as in \cite{huang2017multi}, and evaluate the validation accuracy. The training setup for ImageNet classification is provided in \cref{sec:training_settings}. On Something-Something V1, we replace the CNN backbone in TSM \cite{lin2019tsm} with ours and follow all the data-processing setups in \cite{lin2019tsm}. In COCO object detection, ImageNet-pretrained models are finetuned for 12 epochs with the default configuration of RetinaNet \cite{lin2017focal} in MMDetection \cite{chen2019mmdetection}.

\subsection{Main results} \label{sec:main_results}

\noindent\textbf{Results on ResNet} are shown in \cref{fig:main_results} (a). The early-exiting performance of our Dyn-Perceiver is represented in gray curves, with the highest accuracy under each budget depicted by black curves. The multiple curves correspond to different models, whose detailed configurations are provided in \cref{sec:model_configurations}. We control the model complexity by manipulating the width (0.375-0.75$\times$) of ResNet-50 \cite{he2016resnet} and the number of initial tokens $L$ in the latent code. Our models are compared with various ResNet-based adaptive inference competitors, including layer skipping (Conv-AIG \cite{veit2018convolutional} and SkipNet \cite{wang2018skipnet}), channel skipping (BAS-ResNet \cite{bejnordi2019batch} and Channel Selection \cite{herrmann2018end}), and spatial-wise dynamic networks (DynConv \cite{verelst_dynamic_2020} and LASNet \cite{han2022latency}).  It can be observed that Dyn-Perceiver significantly outperforms other types of dynamic networks. Notably, apart from the performance, a key advantage of Dyn-Perceiver is its flexibility to adjust the computational cost with \emph{a single} model. When the resource budget varies, we can simply set appropriate early-exiting thresholds to meet the constraint instead of training another model with different sparsity like other methods.

\noindent\textbf{Results on RegNets.} 
We further implement Dyn-Perceiver on RegNet-Y \cite{radosavovic2020designing} from 400M to 3.2G FLOPs and compare our method with multiple \emph{static} backbones. The results in \cref{fig:main_results} (b) show the consistent improvement of our method across a wide range of computational budgets. For instance, Dyn-Perceiver reduces the computation by $4.8\times$ to achieve the same accuracy as a RegNet-Y-4GF. 
Compared with the recent Swin-Transformer \cite{liu2021swin} and Vision Transformer with Deformable Attention (DAT) \cite{xia2022vision}, Dyn-Perceiver reduces the computation by $1.8\times$ and $1.4\times$ respectively.

\noindent\textbf{Comparison with early-exiting networks.} 
Our RegNet-based Dyn-Perceiver is also compared with state-of-the-art dynamic early-exiting networks, including MSDNet \cite{huang2017multi}, RANet \cite{yang2020resolution}, MSDNet trained with improved training techniques \cite{li2019improved,han2022learning}, glance-and-focus network (GFNet) \cite{wang2020glance}, dynamic vision Transformer (DVT) \cite{wang2021images}, and the recent CF-ViT \cite{chen2022coarse}. The results in \cref{fig:main_results} (c) demonstrate that Dyn-Perceiver consistently outperforms these competitors.


\noindent\textbf{Results on MobileNet-v3.} We further validate Dyn-Perceiver on MobileNet-v3 with different width factors (0.75-1.5$\times$). Our method is compared with various competitive baselines, including CNN (MobileNet-v3 \cite{howard2019searching}), vision Transformers (DeiT \cite{touvron2021training}, T2T-ViT \cite{yuan2021tokens}, PVT \cite{wang2021pyramid}), and hybrid models (MobileViT \cite{mehta2022mobilevit}, MobleViT-v2 \cite{mehta2022separable}, EfficientFormer \cite{li2022efficientformer}, EdgeViT \cite{pan2022edgevits}, Lite Vision Transformer (LVT) \cite{yang2022lite}, EdgeNeXt \cite{maaz2022edgenext} and MixFormer \cite{chen2022mixformer}). As can be seen from  \cref{fig8_mob_results} that Dyn-Perceiver consistently outperforms the competitors in a wide range of computational budgets. For example, when the budget ranges in 0.2-0.8 GFLOPs, Dyn-Perceiver has $\sim\!\!$ 1.2-1.4$\times$ less computation than MobileNet-v3 when achieving the same performance. 

\noindent\textbf{Action recognition.} We implement ResNet-based (for a fair comparison with baselines) Dyn-Perceiver in the TSM framework \cite{lin2019tsm} and compare
our method with competitors including TSM \cite{lin2019tsm}, TRN \cite{zhou2018temporal}, ECO \cite{zolfaghari2018eco} and AdaFuse \cite{meng2021adafuse}. As shown in \cref{fig_video_results}, Dyn-Perceiver can seamlessly be applied in video classification and achieves a favorable trade-off between accuracy and efficiency.

\begin{figure}
        \centering
        \includegraphics[width=\linewidth]{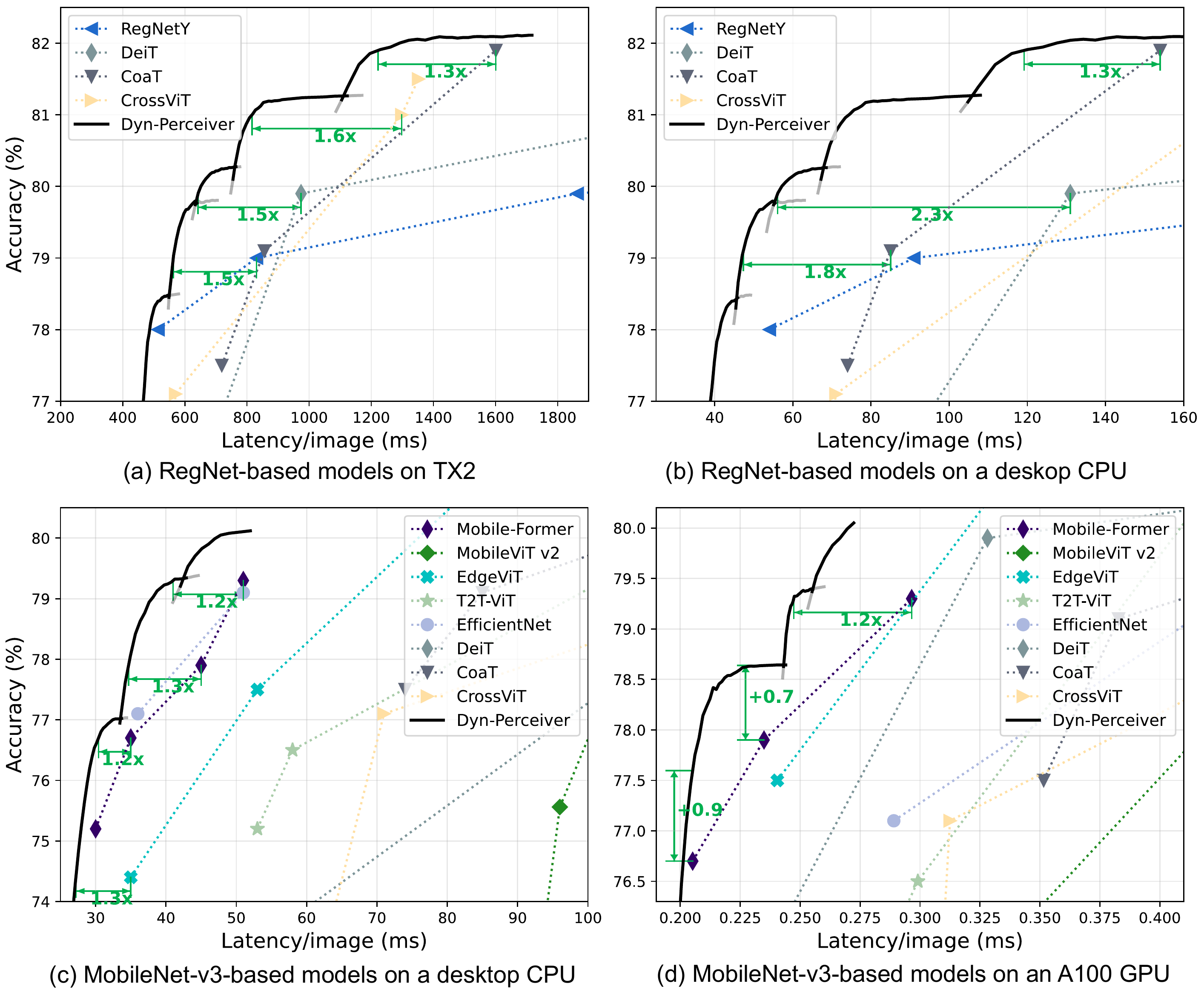}
    \vskip -0.1in
    \caption{\textbf{Speed test results of Dyn-Perceiver} on different hardware platforms including a mobile device TX2 (a), a desktop CPU (b, c), and a server-end A100 GPU (d).}
    \label{fig:latency_dyn_vs_mobileformer}
    \vskip -0.1in
\end{figure}

\begin{table}
  \begin{center}
  \tablestyle{3pt}{1.1}
  \resizebox{\linewidth}{!}{
  \begin{tabular}{c c c c c c}
  \multirow{2}{*}{Model} & Key/Value & $\mathbf{Z2X}$ /  & \multirow{2}{*}{Dyn} & \multirow{2}{*}{FLOPs} & Top-1 \\[-0.5ex]
  &  in $\mathbf{X2Z}$ & $\mathbf{Z2O}$ &  &  & Acc. (\%)   \\
  \shline
  Perceiver \cite{pmlr-v139-jaegle21a} & Input image & \xmark & \xmark & 404G & 78.6 \\
  Perceiver IO \cite{jaegle2021perceiver} & Input image & \cmark & \xmark & 407G & 79.0 \\
  \cellcolor{lightgray!50}\multirow{1}{*}{Dyn-Perceiver} & \cellcolor{lightgray!50}Feature map & \cellcolor{lightgray!50}\cmark & \cellcolor{lightgray!50}\xmark & \cellcolor{lightgray!50}{0.86G} & \cellcolor{lightgray!50}\textbf{79.5} \\[-0.5ex] 
  \cellcolor{lightgray!50}(MobileNet-v3-1.25$\times$) & \cellcolor{lightgray!50}Feature map & \cellcolor{lightgray!50}\cmark & \cellcolor{lightgray!50}\cmark & \cellcolor{lightgray!50}\textbf{0.55G} & \cellcolor{lightgray!50}{79.0} \\ 
  \end{tabular}
    }
    \vskip -0.1in
    \caption{\textbf{Dyn-Perceiver \emph{v.s.} Perceiver}  \cite{pmlr-v139-jaegle21a} \textbf{and Perciver IO} \cite{jaegle2021perceiver}. All three models use a latent code $\mathbf{Z}$ to query information from input images \cite{pmlr-v139-jaegle21a,jaegle2021perceiver} or features $\mathbf{X}$ (ours). Perceiver IO \cite{jaegle2021perceiver} adopts an additional output code $\mathbf{O}$ for classification. ``Dyn'' represents dynamic early exiting.}
    \label{tab:comparison_with_perceiver}
  \end{center}
  \vskip -0.3in
\end{table} 

\noindent\textbf{The practical efficiency.} We evaluate the practical efficiency of Dyn-Perceiver across multiple hardware platforms, including a mobile device (Nvidia Jetson TX2), a desktop CPU (Intel i5-8265U), and a server GPU (Nvidia A100). We set the batch size to 1 on CPUs and 128 on GPU. The accuracy-latency curves in \cref{fig:latency_dyn_vs_mobileformer} demonstrate that the exceptional \emph{theoretical} efficiency of dynamic early exiting can effectively translate into the \emph{realistic speedup} across different hardware platforms. For instance, while the recent Mobile-Former \cite{chen2022mobile} exhibits remarkable performance in theoretical efficiency, the MobileNet-v3-based Dyn-Perceiver models consistently run faster on hardware while achieving comparable accuracy. We conjecture this is because our two-branch structure can be executed in parallel, and the regular activation functions are more hardware-friendly compared to dynamic ReLU \cite{chen_dynamic_2020_relu} adopted in \cite{chen2022mobile}.


\begin{table}
  \vskip -0.05in
  \begin{center}
  \tablestyle{8pt}{1.1}
  \begin{tabular}{l c l}
  Exit  & 1 & 2 \\
  \shline
 \multicolumn{2}{l}{\textbf{\emph{CNN}} (0.86 GFLOPs)} & \\
  \multicolumn{1}{l}{RegNet-Y-800MF \cite{radosavovic2020designing} w/o EE} & - & 77.0 \\
  ~~~~w/ EE  & 72.4 & 75.8 \scriptsize{({\color{gray}{$\downarrow$1.2}})} \\
  \hline
  \multicolumn{2}{l}{\textbf{\emph{Transformer w/ CLS token}} (1.10 GFLOPs)}  \\
  \multicolumn{1}{l}{T2T-ViT-7 \cite{yuan2021tokens} w/o EE}  & - & 71.7 \\
  ~~~~w/ EE & 65.4  & 69.5 \scriptsize{({\color{gray}{$\downarrow$}2.2})}\\
  \hline
  \textbf{\emph{Dyn-Perceiver}} (0.67 GFLOPs)  \\
  w/o EE & - & 77.1 \\
  ~~~~w/ EE in the feature branch  & 75.6 & 76.9 \scriptsize{({\color{gray}{$\downarrow$0.2}})} \\
  \cellcolor{lightgray!50}~~~~w/ EE in the classification branch &  \cellcolor{lightgray!50}\textbf{76.8} & \cellcolor{lightgray!50}\textbf{77.8} \scriptsize{({\color{blue}{$\uparrow$0.7}})}  \\
  \end{tabular}
    \vskip -0.1in
    \caption{\textbf{Ablation study of the two-branch structure}. The results on RegNet and T2T-ViT show the negative effect when an early classifier (EE) is built on intermediate features or CLS token. In contrast, the EE in our classification branch \emph{increases} the last classifier's accuracy.}
    \label{tab:ablation_cls}
  \end{center}
  \vskip -0.3in
\end{table}

\noindent\textbf{Comparison with Perceiver} \cite{pmlr-v139-jaegle21a} \textbf{and Perceiver IO} \cite{jaegle2021perceiver} is presented in \cref{tab:comparison_with_perceiver}. By introducing the \emph{feature branch}, the symmetric cross-attention mechanism, and the dynamic early-exiting paradigm, our method significantly reduces the computation without sacrificing accuracy.

\subsection{Ablation studies} \label{sec:ablation}
We conduct ablation studies with our RegNetY-400MF-based model to validate the effectiveness of our two-branch framework and different design choices.

\noindent\textbf{The effectiveness of our two-branch framework.} We first demonstrate that incorporating early exits (EE) into a standard model degrades its final performance. 
We experiment with a CNN (RegNet-Y-800M \cite{radosavovic2020designing}) and a vision Transformer with a CLS token (T2T-ViT-7 \cite{yuan2021tokens}). An early exit is built on the feature map at stage 3 of the CNN and the CLS token at the 4$^\mathrm{th}$ block of T2T-ViT-7, respectively. The accuracy of different exits is reported in \cref{tab:ablation_cls}. It is observed that the performance of both the CNN and the vision Transformer is significantly affected by the early exit. We can conclude that the CLS token cannot serve as our latent code perfectly, as it frequently participates in the attention operation with other patches in each block, and the network weights for processing the CLS token are shared with those for processing image features. In other words, feature extraction and early classification are still closely coupled.

Next, we conduct experiments with our two-branch structure, which is compared with two variants: the first has the same architecture but without early exits (EE), and the second incorporates an EE in the feature branch. The accuracy of different exits is listed in \cref{tab:ablation_cls}. The results suggest that: 1) the \emph{classification branch} mitigates the accuracy drop brought by the EE to some extent, even if it is placed in the feature branch; 2) the EE in the feature branch still downgrades the last classifier's performance by interfering with feature extraction; 3) \emph{building EE in the classification branch} successfully decouples feature extraction and early classification and is therefore superior to the former choice. Moreover, the last exit even \emph{outperforms the variant without any EE}. We conjecture that the EE provides a ``deep supervision'' \cite{lee2015deeply} for the classification branch. The above analysis indicates that our two-branch architecture is the key to avoiding the negative effect brought by early exits.

\begin{table}
  \begin{center}
  \tablestyle{4pt}{1.1}
  \resizebox{\linewidth}{!}{
  \begin{tabular}{l | l l l l}
  & \multicolumn{4}{c}{Top-1 Acc. (\%)} \\[-0.5ex]
  Exit index  &1 & 2 & 3 & 4 \\
  \shline
  Vanilla & 43.0 &  73.6 & 50.1 & 74.3 \\
  + $\mathbf{X2Z}$ & 57.3 \scriptsize{(\textcolor{blue}{$\uparrow 14.3$})} & 73.4 \scriptsize{(\textcolor{gray}{$\downarrow 0.2$})} & 74.8 \scriptsize{(\textcolor{blue}{$\uparrow 24.7$})} & 76.5 \scriptsize{(\textcolor{blue}{$\uparrow 2.2$})} \\
  + DWC & 58.0 \scriptsize{(\textcolor{blue}{$\uparrow 0.7$})} &  73.6 \scriptsize{(\textcolor{blue}{$\uparrow 0.2$})} & 75.0 \scriptsize{(\textcolor{blue}{$\uparrow 0.2$})} & 76.7 \scriptsize{(\textcolor{blue}{$\uparrow 0.2$})} \\
  + $\mathbf{Z2X}$  & 56.3 \scriptsize{(\textcolor{gray}{$\downarrow 1.7$})} & 75.1 \scriptsize{(\textcolor{blue}{$\uparrow 1.5$})} & 75.9 \scriptsize{(\textcolor{blue}{$\uparrow 0.9$})} & 77.8 \scriptsize{(\textcolor{blue}{$\uparrow 1.1$})} \\
  + FKT  & 57.3 \scriptsize{(\textcolor{blue}{$\uparrow 1.0$})}  & 75.4 \scriptsize{(\textcolor{blue}{$\uparrow 0.3$})} & 76.2 \scriptsize{(\textcolor{blue}{$\uparrow 0.3$})} & \textbf{77.9} \scriptsize{(\textcolor{blue}{$\uparrow 0.1$})} \\
  \cellcolor{lightgray!50}+ Self-distillation  & \cellcolor{lightgray!50}\textbf{61.8 \scriptsize{(\textcolor{blue}{$\uparrow 4.5$})}} &  \cellcolor{lightgray!50}\textbf{75.6 \scriptsize{(\textcolor{blue}{$\uparrow 0.2$})}} & \cellcolor{lightgray!50}\textbf{76.8 \scriptsize{(\textcolor{blue}{$\uparrow 0.6$})}} & \cellcolor{lightgray!50}77.8 \scriptsize{(\textcolor{gray}{$\downarrow 0.1$})} \\
  - Token Mixers & 42.3 \scriptsize{(\textcolor{gray}{$\downarrow 19.5$})} & 74.9 \scriptsize{(\textcolor{gray}{$\downarrow 0.7$})} & 59.1 \scriptsize{(\textcolor{gray}{$\downarrow 17.7$})}  & 75.2 \scriptsize{(\textcolor{gray}{$\downarrow 2.6$})} \\
  \end{tabular}
    }
    \vskip -0.1in
    \caption{\textbf{Ablation study of different components}. Each row increases neglectable computation overhead.}
    \label{tab:ablation_main}
  \vskip -0.3in
  \end{center}
\end{table}
\begin{figure}
        \vskip -0.2in
        \centering
        \includegraphics[width=\linewidth]{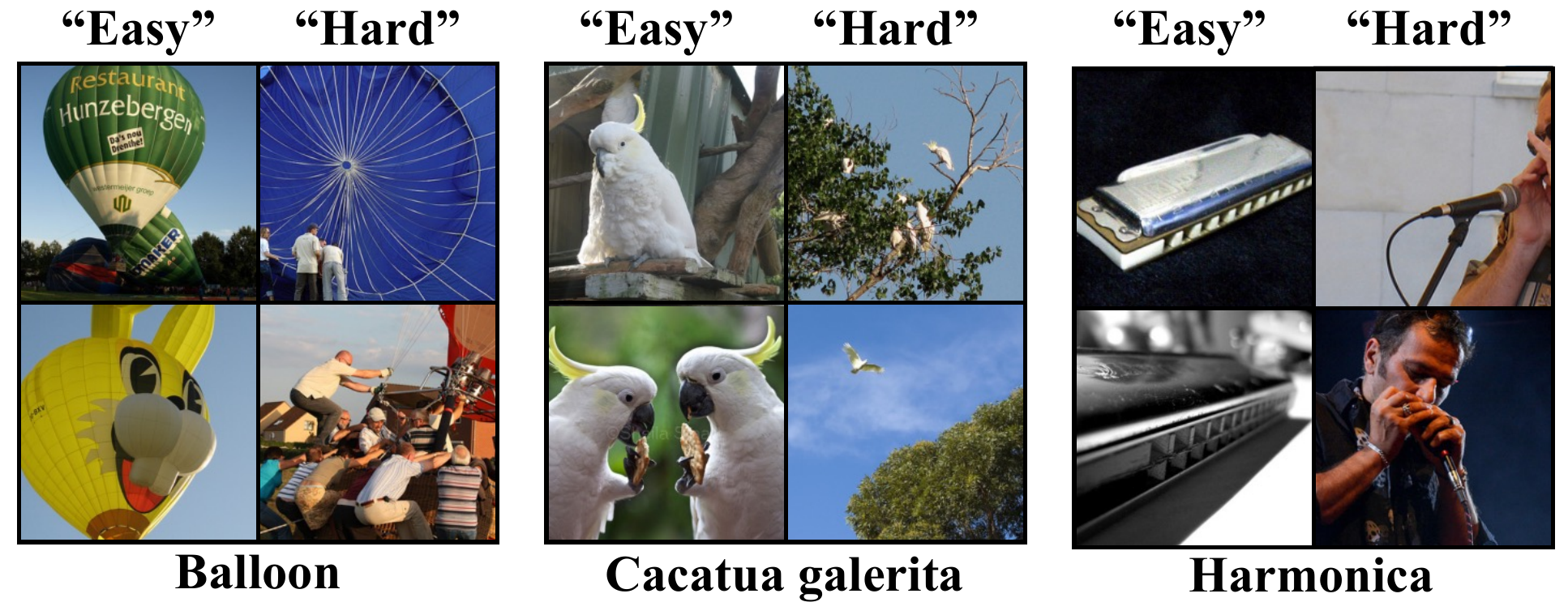}
        \vskip -0.1in
        \caption{Visualization of ``easy'' and ``hard'' samples.}
        \label{fig_visualization}
    \vskip -0.2in
\end{figure}
\noindent\textbf{The effect of different components.} We start from a ``vanilla'' two-branch structure that lacks cross-attention layers and FKT modules, and train it without self-distillation. In the vanilla model, we retain the first $\mathbf{X2Z}$ cross attention, as we find the training divergent without it. Then we progressively add the components introduced in \cref{sec_main_components}. The first line in \cref{tab:ablation_main} shows that the classification branch performs subpar without aggregating sufficient information from the feature branch. 
Next, the $\mathbf{X2Z}$ cross attention significantly improves the performance of the classification branch. The 4$^\mathrm{th}$ line of \cref{tab:ablation_main} demonstrates that the semantic information in the latent code also bolsters the performance of the feature branch. FKT and self-distillation further improve the accuracy of early classifiers. Finally, we remove the token mixers, which means that the token length and the channel number of the latent code are kept the same across different stages. We can witness that the token mixers are also important to the final performance.

\subsection{Visualization} \label{sec:visualization}
\cref{fig_visualization} show the images that are output by the first (``easy'') and the last (``hard'') exit of Dyn-Perceiver during dynamic early exiting. We can easily tell that ``easy'' samples generally contain simpler backgrounds, and the foreground objects usually have clearer appearances and standard poses. In the ``hard'' images, the foreground objects may have incomplete appearances (\emph{e.g.} the balloon) or are very small in the scene (\emph{e.g.} the Cacatua galerita). Interestingly, the harmonicas don't even appear in the hard images, yet they are still correctly classified by the last exit. This indicates that our latent code captures rich semantic-level information to understand the ``playing harmonica'' action.

\begin{table}
  \begin{center}
  \tablestyle{2pt}{1.05}
  \begin{tabular}{c c c c c c c}
  \multirow{2}{*}{Model} & Backbone &  \multirow{2}{*}{mAP} & \multirow{2}{*}{mAP$^{{s}}$} & \multirow{2}{*}{mAP$^{{m}}$} & \multirow{2}{*}{mAP$^{{l}}$} \\[-0.5ex]
   &  FLOPs &   \\
  \shline
   RegNet-X-1.6GF  & 33.2G & 37.4 & 22.4 & 41.1 & 49.2  \\
   RegNet-Y-1.6GF* & 33.4G & 38.5 & 22.1 & 41.7 & 50.8  \\
   RegNet-X-3.2GF & 65.5G & 39.0 & 22.6 & 43.5 & 50.8 \\
   RegNet-Y-3.2GF* & 66.0G & 39.3 & 22.7 & 43.1 & 51.8  \\
   
   \cellcolor{lightgray!50}Dyn-Perceiver  & \cellcolor{lightgray!50} & \cellcolor{lightgray!50} & \cellcolor{lightgray!50} & \cellcolor{lightgray!50} & \cellcolor{lightgray!50}  \\[-0.5ex]
   \cellcolor{lightgray!50}(RegNet-Y-1.6GF) & \cellcolor{lightgray!50}\multirow{-2}{*}{\textbf{38.1G}} & \cellcolor{lightgray!50}\multirow{-2}{*}{\textbf{40.2}} & \cellcolor{lightgray!50}\multirow{-2}{*}{\textbf{24.1}} & \cellcolor{lightgray!50}\multirow{-2}{*}{\textbf{43.9}} & \cellcolor{lightgray!50}\multirow{-2}{*}{\textbf{53.0}} \\
  \end{tabular}
    \vskip -0.1in
    \caption{\textbf{COCO detection results}. The RegNet-X performance is obtained from the official website, and RegNet-Y* is our implementation. The metrics mAP$^{{s}}$, mAP$^{{m}}$, and mAP$^{{l}}$ denote the mAP on small, medium, and large objects.}
    \label{tab:coco_det}
  \end{center}
  \vskip -0.3in
\end{table}

\subsection{Object detection results} \label{sec:coco_det}
The recent early-exiting networks \cite{huang2017multi,yang2020resolution,wang2021images} usually have specially designed architectures, and may not be suitable to apply on downstream tasks, \emph{e.g.} object detection. In contrast, Dyn-Perceiver can be built on top of standard vision models, and therefore can seamlessly serve as a backbone for object detection. We implement a RegNet-based model in RetinaNet \cite{lin2017focal}. Mean average precision (mAP) on the COCO \cite{lin2014microsoft} validation set is used to measure the detection performance. Note that early exiting is not used in this task, and the experiment is mainly to demonstrate the generality of our model. The results in \cref{tab:coco_det} suggest that Dyn-Perceiver outperforms the baselines even with less computation. The performance on object detection further validates that \emph{early classifiers in the classification branch will not downgrade the quality of the feature pyramid extracted by the feature branch}. To the best of our knowledge, Dyn-Perceiver is the first dynamic early-exiting network that is empirically evaluated in the object detection task.
\section{Conclusion} \label{sec:conclusion}
We introduced Dynamic Perceiver (Dyn-Perceiver), to explicitly decouple feature extraction and early exiting with a two-branch structure for efficient visual recognition. The inspiration came from the general-purpose architecture Perceiver using a latent code to directly query inputs. We adapted Perceiver to efficient visual recognition by introducing a feature branch. The latent code in Dyn-Perceiver is processed by a classification branch, and early exits are only inserted in this classification branch, thus not affecting the coarse-to-fine feature extraction process. The two branches interact with each other via symmetric cross-attention layers. Experiments on ImageNet demonstrated that our design effectively mitigates the negative effects brought by early exits. The framework consistently reduced the computational cost of different visual backbones on image classification, action recognition, and object detection tasks. Dyn-Perceiver significantly outperformed various competitive models in balancing accuracy and efficiency. Furthermore, evaluations on multiple hardware devices showcased the preferable inference latency of our method.

\noindent\textbf{Acknowledgement.} This work is supported in part by National Key R\&D Program of China (2021ZD0140407), the National Natural Science Foundation of China (62022048, 62276150) and the Tsinghua University-China Mobile Communications Group Co.,Ltd. Joint Institute. We also appreciate the generous donation of computing resources by High-Flyer AI.

{\small
\bibliographystyle{ieee_fullname}
\bibliography{ref}
}

\clearpage
\section*{Appendix}
\appendix
We provide more details of our experiments, including the configuration of all models reported in the main text (\cref{sec:model_configurations}) and the training setup (\cref{sec:training_settings}). We also provide the speed test results of Dyn-Perceiver implemented on ResNet (\cref{sec:speed_resnet}).

\section{Model Configuration} \label{sec:model_configurations}
In \cref{fig:main_results} and \cref{fig8_mob_results}, we reported the results of different-sized Dyn-Perceiver. Here we list the detailed configuration of Dyn-Perceiver implemented on ResNet (\cref{tab:model_config_res}), RegNet-Y (\cref{tab:model_config_reg}) and MobileNet-v3 (\cref{tab:model_config_mob}). SA is the abbreviation of self-attention. 

\section{Training Settings} \label{sec:training_settings}
\noindent\textbf{Image classification.} The training settings of Dyn-Perceiver are demonstrated in \cref{tab:training_settings}. For simplicity, we use the same settings to train MobileNet-based models and the ResNet/RegNet-based models, except for the training epochs. For each experiment, we select batch size from \{1024, 2048\} based on model sizes and the GPU memory.

\noindent\textbf{Action recognition.} We use the TSM \cite{lin2019tsm} code base\footnote{\url{https://github.com/mit-han-lab/temporal-shift-module}.} and add temporal shift to our latent code $\mathbf{Z}$ in self-attention blocks. We follow the settings in the official implementation and sum up the classification loss from different exits with the same weights as in ImageNet training.

\noindent\textbf{Object detection.} We finetune the ImageNet-pretrained checkpoints on COCO for 12 epochs following the official settings of RetinaNet \cite{lin2017focal} in the MMDetection code base\footnote{\url{https://www.github.com/open-mmlab/mmdetection}.} Since the feature maps from different stages are required by the detection head, we do not perform early exiting here, and the experiment is only to demonstrate the capability of Dyn-Perceiver to serve as a detection backbone.

\section{Speed Test for ResNet-based Model} \label{sec:speed_resnet}
\begin{figure}
    \begin{subfigure}[t]{0.85\linewidth}
        \centering
        \includegraphics[width=\linewidth]{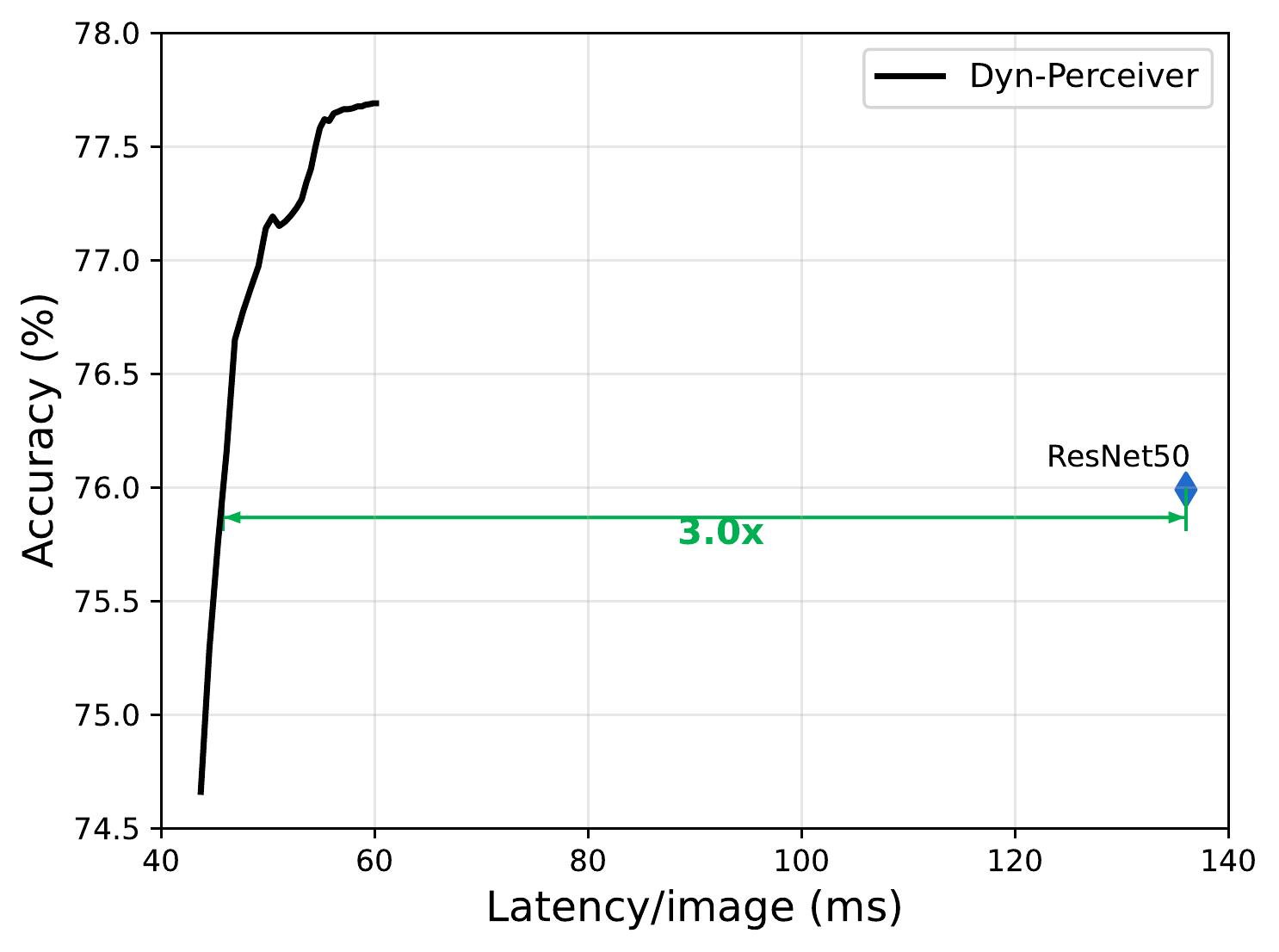}
        \caption{CPU.}
        \label{fig_res_cpu}
    \end{subfigure}\hfill
    \begin{subfigure}[t]{0.85\linewidth}
        \centering
        \includegraphics[width=\textwidth]{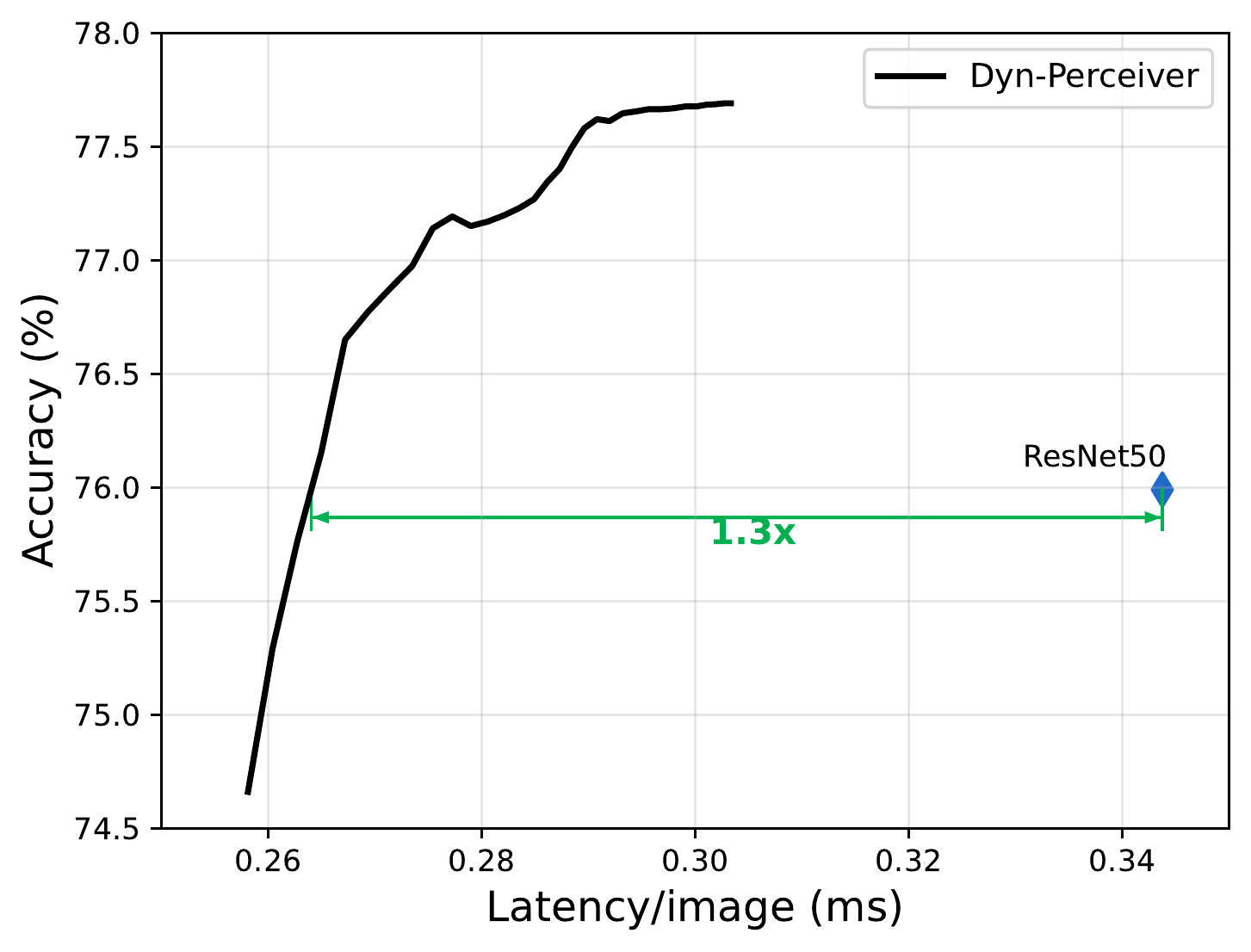}
        \caption{GPU.}
        \label{fig_res_gpu}
    \end{subfigure}
    \caption{Speed test results of Dyn-Perceiver built on ResNet (0.375$\times$, model 1 in \cref{tab:model_config_res}).}
\end{figure}

We also test the practical efficiency of our smallest ResNet-based Dyn-Perceiver (model 1 in \cref{tab:model_config_res}) on the desktop i5 CPU and the A100 GPU. The latency-accuracy curves on CPU and GPU are presented in \cref{fig_res_cpu} and \cref{fig_res_gpu}, respectively. It could be observed that Dyn-Perceiver achieves satisfying speedup on the two devices when achieving the same accuracy with the baseline.

\begin{table}
  \begin{center}
  \tablestyle{2pt}{1.2}
     \resizebox{\linewidth}{!}{
  \begin{tabular}{l | c c c c c}
  & \multicolumn{5}{c}{ResNet50-based Model} \\
  Model index & 1 & 2 & 3 & 4 & 5 \\
  \shline
  ResNet width factor & 0.375$\times$ & 0.5$\times$ & 0.5$\times$ & 0.625$\times$ & 0.75$\times$ \\
  token number $L$ & 128 & 128 & 256 & 192 & 128\\
  \# SA blocks & [3,3,9,3] & [3,3,9,9] & [3,3,9,3] & [3,3,9,3] & [3,3,9,3] \\
  SA widening factor  & 4 & 4 & 4 & 4 & 2 \\
  \end{tabular}
  }
    \caption{ResNet-50 configurations. The 5 configurations correspond to the 5 curves in \cref{fig:main_results} (a) of the paper.}
    \label{tab:model_config_res}
  \end{center}
\end{table}

\begin{table}
  \begin{center}
  \tablestyle{2pt}{1.2}
    \vskip -0.1in
     \resizebox{\linewidth}{!}{
  \begin{tabular}{l | c c c c c c}
  & \multicolumn{6}{c}{RegNet-based Model} \\
  Model index & 1 & 2 & 3 & 4 & 5 & 6 \\
  \shline
  RegNet size & 400M & 400M & 800M & 800M & 1.6G & 3.2G \\
  token number $L$ & 128 & 256 & 128 & 256 & 256 & 256\\
  \# SA blocks & [6,6,9,9] & [3,3,9,9] & [3,3,9,6] & [3,3,9,6] & [6,6,9,6] & [6,6,9,9] \\
  SA widening factor  & 4 & 4 & 4 & 4 & 2 & 4 \\
  \end{tabular}
  }
    \caption{RegNet-Y configurations. The 6 configurations correspond to the 6 curves in \cref{fig:main_results} (b) of the paper.}
    \label{tab:model_config_reg}
  \end{center}
\end{table}

\begin{table}
  \begin{center}
  \tablestyle{2pt}{1.2}
  \begin{tabular}{l | c c c c c}
  & \multicolumn{5}{c}{MobileNet-based Model} \\
  Model index & 1 & 2 & 3 & 4 & 5 \\
  \shline
  MobileNet width factor & 0.75× & 1.00× & 1.00× & 1.25× & 1.5× \\
  token number $L$ & 128 & 128 & 128 & 128 & 256 \\
  \# SA blocks & [3,3,9,9] & [3,3,9,9] & [6,6,9,9] & [6,6,9,9] & [3,3,9,9] \\
  SA widening factor  & 4 & 4 & 4 & 4 & 4 \\
  \end{tabular}
    \caption{MobileNet-v3 configurations. The 5 configurations correspond to the 5 curves in \cref{fig8_mob_results} of the paper.}
    \label{tab:model_config_mob}
  \end{center}
\end{table}

\begin{table}
  \begin{center}
  \tablestyle{4pt}{1.2}
    \resizebox{\linewidth}{!}{
  \begin{tabular}{l | c c}
  \multirow{2}{*}{Training Config} & Dyn-Perceiver & Dyn-Perceiver \\[-0.5ex]
   & MobileNet-based & ResNet/RegNet-based  \\
  \shline
   optimizer  & adamW & adamW \\
   base learning rate  & \{1e-3, 2e-3\} & \{1e-3, 2e-3\} \\
   weight decay  & 0.05 & 0.05 \\
   optimizer momentum  & $\beta_{1}$, $\beta_{2}$=0.9, 0.999 & $\beta_{1}$, $\beta_{2}$=0.9, 0.999 \\
   batch size  & \{1024, 2048\} & \{1024, 2048\} \\
   training epochs  & 600 & 300 \\
   learning rate schedule  & cosine decay & cosine decay \\
   warmup epochs  & 20 & 20 \\
   warmup schedule  & linear & linear \\
   layer-wise lr decay \cite{bao2021beit} & None & None \\
   randaugment \cite{cubuk2020randaugment} & (9, 0.5) & (9, 0.5) \\
   label smoothing \cite{szegedy2016rethinking} & 0.1 & 0.1 \\
   mixup \cite{zhang2018mixup} & 0.8 & 0.8 \\
   cutmix \cite{yun2019cutmix} & 1.0 & 1.0 \\
   stochastic depth \cite{huang2016deep} & None & None \\
   layer scale \cite{touvron2021going} & None & None \\
   gradient clip  & None & None \\
   exp. mov. agv.(EMA) \cite{polyak1992acceleration}  & 0.9999 & 0.9999 \\
  \end{tabular}
  }
    \caption{ImageNet-1K training settings.}
    \label{tab:training_settings}
  \end{center}
\end{table}

\end{document}